\documentclass[journal]{IEEEtran}
 \pdfoutput=1
\usepackage{scalerel}
\usepackage{tikz}
\usetikzlibrary{svg.path}
\definecolor{pumpkin}{RGB}{211, 84, 0}
\definecolor{orcidlogocol}{HTML}{A6CE39}
\tikzset{
  orcidlogo/.pic={
    \fill[orcidlogocol] svg{M256,128c0,70.7-57.3,128-128,128C57.3,256,0,198.7,0,128C0,57.3,57.3,0,128,0C198.7,0,256,57.3,256,128z};
    \fill[white] svg{M86.3,186.2H70.9V79.1h15.4v48.4V186.2z}
                 svg{M108.9,79.1h41.6c39.6,0,57,28.3,57,53.6c0,27.5-21.5,53.6-56.8,53.6h-41.8V79.1z M124.3,172.4h24.5c34.9,0,42.9-26.5,42.9-39.7c0-21.5-13.7-39.7-43.7-39.7h-23.7V172.4z}
                 svg{M88.7,56.8c0,5.5-4.5,10.1-10.1,10.1c-5.6,0-10.1-4.6-10.1-10.1c0-5.6,4.5-10.1,10.1-10.1C84.2,46.7,88.7,51.3,88.7,56.8z};
  }
}

\newcommand\orcidicon[1]{\href{https://orcid.org/#1}{\mbox{\scalerel*{
\begin{tikzpicture}[yscale=-1,transform shape]
\pic{orcidlogo};
\end{tikzpicture}
}{|}}}}

\usepackage[utf8]{inputenc}
\usepackage[T1]{fontenc}
\usepackage{microtype,newtxtext,newtxmath} 
\usepackage{mathtools,gensymb,eurosym,url} 
\usepackage{caption}
\usepackage{subcaption}
\usepackage{longtable,booktabs,array}
\usepackage{colortbl,multirow,tabularx}
\usepackage{verbatim}
\usepackage{soul,xcolor}
\usepackage{graphicx}
\usepackage[backend=biber, sorting=none, style=numeric-comp, maxnames=3]{biblatex}
\usepackage{acronym}
\usepackage{hyperref}
\usepackage{algpseudocode}
\hypersetup{colorlinks=true, breaklinks=true, 
  urlcolor=blue, linkcolor=blue,anchorcolor=blue,citecolor=blue,
  hypertexnames=true, final=true,
  pdfpagemode = UseNone, 
  pdfauthor = {},
  pdftitle = {},
  pdfsubject = {},
  pdfkeywords = {}
}

\graphicspath{{figures/}}
\DeclareGraphicsExtensions{.png,.pdf}
\DeclareSourcemap{
  \maps{
    \map{
      \step[fieldset=month, null]
      \step[fieldset=publisher,null]
      \step[fieldset=address, null]
      \step[fieldset=location, null]
      \step[fieldset=doi, null]
      \step[fieldset=url, null]
      \step[fieldset=isbn, null]
      \step[fieldset=number, null]
      \step[fieldset=pages, null]
      \step[fieldset=note, null]
      \step[fieldset=urldate, null]
      \step[fieldset=language,null]
      \step[fieldset=issn_a,null]
      \step[fieldset=issn, null]
    }
  }
}
\renewbibmacro{in:}{}

\acrodef{ADM}[ADM]{asynchronous  delta modulator}
\acrodef{AdaBN}[AdaBN]{Adaptive Batch Normalization}
\acrodef{BN}[BN]{Batch Normalization}
\acrodef{CL}[CL]{Continual Learning} 
\acrodef{SoTA}[SoTA]{state-of-the-art}
\acrodef{CNN}[CNN]{Convolutional Neural Network} 
\acrodef{Cov}[Cov]{Contrasts covariance matching}
\acrodef{DA}[DA]{Domain Adaptation} 
\acrodef{DER}[DER]{Dark Experience Replay} 
\acrodef{ECG}[ECG]{electrocardiography}
\acrodef{EMG}[EMG]{electromyography}
\acrodef{EEG}[EEG]{electroencephalography}
\acrodef{ECoG}[ECoG]{electrocorticography}
\acrodef{EMA}[EMA]{Exponential Moving Average} 
\acrodef{ENG}[ENG]{electroneurographic}
\acrodef{GMM}[GMM]{Gaussian Mixture Model}
\acrodef{HFO}[HFO]{High-Frequency Oscillations}
\acrodef{LIF}[LIF]{Leaky Integrate and Fire}
\acrodef{LIFE}[LIFE]{Longitudinal Intrafascicular Electrode}
\acrodef{TIME}[TIME]{Transverse Intrafascicular Multichannel Electrode}
\acrodef{tfLIFE}[tf-LIFE]{thin-film LIFE} 
\acrodef{DoF}[DoF]{degree-of-freedom}
\acrodefplural{DoF}{degrees-of-freedom}
\acrodef{LoRA}[LoRA]{Low-Rank Adaptation}
\acrodef{LR}[LR]{Latent Replay} 
\acrodef{MMAC}[MMAC]{multiply–accumulate operations} 
\acrodef{MAV}[MAV]{mean absolute value} 
\acrodef{MAML}[MAML]{model-agnostic meta-learning}
\acrodef{SNN}[SNN]{Spiking Neural Network}
\acrodef{USEA}[USEA]{Utah slanted electrode array}
\acrodef{RBF}[RBF]{radial basis function}
\acrodef{RMS}[RMS]{root mean square}
\acrodef{SWD}[SWD]{Sliced Wasserstein Discrepancy}
\acrodef{SVM}[SVM]{support vector machine}
\acrodef{LDA}[LDA]{Linear Discriminant Analysis}
\acrodef{ANN}[ANN]{artificial neural network}
\acrodef{PCA}[PCA]{Principal Component Analysis}
\acrodef{PC}[PC]{Principal Component}
\acrodef{CCA}[CCA]{Canonical Correlation Analysis}
\acrodef{VLSI}[VLSI]{Very Large Scale Integration}
\acrodef{DoF}[DoF]{Degrees of Freedom}
\acrodef{DoA}[DoA]{Degrees of Actuation}
\acrodef{TCN}[TCN]{Temporal Convolutional Network} 
\acrodef{VR}[VR]{virtual reality}
\acrodef{NODE}[NODE]{Neural Ordinary Differential Equations} 
\acrodef{MAE}[MAE]{mean absolute error} 
\acrodef{PCA}[PCA]{Principal Component Analysis}
\acrodef{LoRA}[LoRA]{Low-Rank Adaptation}
\acrodef{MLP}[MLP]{multilayer perceptron}
\acrodef{OOD}[OOD]{out-of-distribution}
\acrodef{UMAP}[UMAP]{Uniform Manifold Approximation and Projection}
\acrodef{SA}[SA]{self-attention}
\acrodef{TL}[TL]{Transfer Learning} 
\acrodef{TTT}[TTT]{Test-Time Training} 
\acrodef{HMI}[HMI]{human-machine interaction}
\acrodef{RF}[RF]{receptive field}

\begin{document}
\title{Lightweight Test-Time Adaptation for EMG-Based Gesture Recognition}
\author{\IEEEauthorblockN{
Nia Touko\IEEEauthorrefmark{1},
Matthew O A Ellis\IEEEauthorrefmark{1},
Cristiano Capone\IEEEauthorrefmark{2},
Alessio Burrello\IEEEauthorrefmark{3},
Elisa Donati\IEEEauthorrefmark{4}\IEEEauthorrefmark{5},
Luca Manneschi\IEEEauthorrefmark{1}\IEEEauthorrefmark{5}}

\IEEEauthorblockA{\IEEEauthorrefmark{1}School of Computer Science, University of Sheffield, Sheffield, S10 2TN, United Kingdom}
\IEEEauthorblockA{\IEEEauthorrefmark{2}Natl.\ Center for Radiation Protection and Computational Physics, Istituto Superiore di Sanità, 00161 Rome, Italy}
\IEEEauthorblockA{\IEEEauthorrefmark{3}Politecnico di Torino, Italy}
\IEEEauthorblockA{\IEEEauthorrefmark{4}Institute of Neuroinformatics, University of Zurich and ETH Zurich, Zurich, Switzerland}
\IEEEauthorblockA{\IEEEauthorrefmark{5}Joint senior authorship}
}
\maketitle

\begin{abstract}
Reliable long-term decoding of surface electromyography (EMG) is hindered by signal drift caused by electrode shifts, muscle fatigue, and posture changes. While state-of-the-art models achieve high intra-session accuracy, their performance often degrades sharply. Existing solutions typically demand large datasets or high-compute pipelines that are impractical for energy-efficient wearables. We propose a lightweight framework for Test-Time Adaptation (TTA) using a Temporal Convolutional Network (TCN) backbone. We introduce three deployment-ready strategies: (i) causal adaptive batch normalization for real-time statistical alignment; (ii) a Gaussian Mixture Model (GMM) alignment with experience replay to prevent forgetting; and (iii) meta-learning for rapid, few-shot calibration. Evaluated on the NinaPro DB6 multi-session dataset, our framework significantly improves generalisation with minimal overhead compared with the non-adaptive baseline. Our results show that test-time adaptation can recover performance lost under out-of-distribution shifts in noisy EMG data, that experience-replay-regularised updates provide superior stability under limited data, and that meta-learning achieves competitive performance in one- and two-shot regimes. Overall, our test-time self-supervised models reach $82\%$ inter-session accuracy, substantially improving upon previous research.  
This work establishes a path toward robust, "plug-and-play" myoelectric control for long-term prosthetic use.
\end{abstract}


\section{Introduction}
\label{sec:intro}
Reliable long-term decoding of neuromuscular signals is the primary bottleneck in developing practical myoelectric interfaces. While surface \ac{EMG} classifiers achieve high accuracy in controlled settings~\cite{Benatti_etal25}, their performance degrades significantly across recording sessions~\cite{Milosevic_etal18}. This decline is driven by the inherent non-stationarity of \ac{EMG} signals, where electrode displacement, muscle fatigue, and changes in skin impedance cause severe distribution shifts, often referred to as signal drift~\cite{Young_etal11, Kaufmann_etal10}. In real-world applications, where daily re-donning of wearable devices is required, models trained on static data often fail to generalize to the target domain.

Early solutions to \ac{EMG} non-stationarity utilized multi-session training or data augmentation to model variability like electrode displacement~\cite{Milosevic_etal18,Kaufmann_etal10}. However, the data intensity and calibration frequency of these methods are incompatible with resource-constrained wearables. Although \ac{TL} and \ac{DA} can improve cross-session robustness by updating a subset of parameters~\cite{CoteAllard2019_adaptation_TL,zhuang_survey_tl} or by performing statistical alignment such as \ac{CCA}~\cite{Donati_etal23}, they typically assume \textit{a priori} access to target-domain data.

Several paradigms offer more flexible alternatives.  \ac{TTT}~\cite{liu2021ttt++} enables online adaptation at deployment by updating model parameters using only incoming (typically unlabelled) data. Depending on the objective, this can range from computationally more demanding self-supervised \cite{he2021autoencoder,gandelsman2022test} training to lightweight variants that adapt only normalization statistics, such as \ac{AdaBN}~\cite{Li_etal16,Ioffe_etal15}. 
To handle non-stationary data streams while mitigating catastrophic forgetting, \ac{CL} proposes a range of strategies, including regularization-based updates and replay-based methods. In particular, replay mechanisms such as \ac{LR}~\cite{ravaglia_lr_riscv} or \ac{DER}~\cite{buzzega2020darkexperiencegeneralcontinual} can maintain performance on prior distributions within tiny memory buffers.
Complementarily, meta-learning optimizes for rapid, few-shot generalization during brief calibration phases. Despite these theoretical frameworks, their application to noisy, high-dimensional \ac{EMG} signals remains largely underexplored.

\begin{figure*}[ht!]
\centering
\includegraphics[width=1.\textwidth]{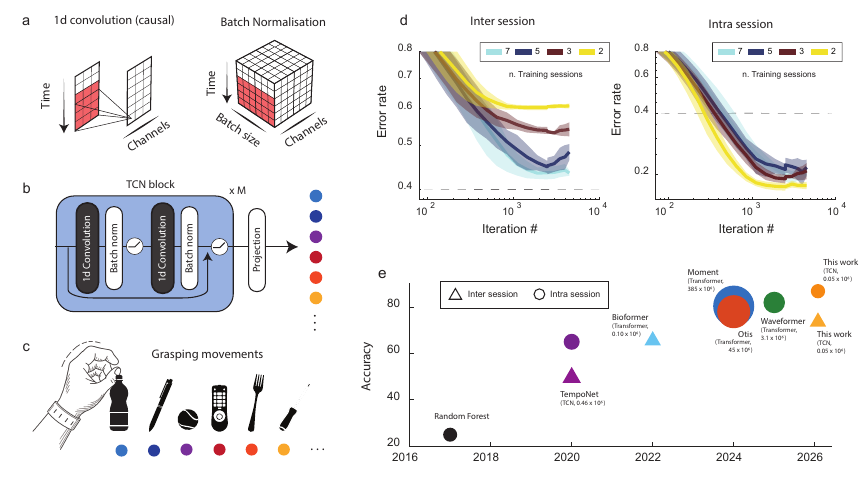}
\caption{Overview of the proposed methodology for inter-session adaptation in surface \ac{EMG} decoding.
(a) 1D causal convolution and batch normalization applied over time, channels, and batch dimensions.
(b) \ac{TCN} architecture with $M$ residual causal-convolution blocks and a linear projection layer.
(c) NinaPro DB6 setup: seven grasping gestures recorded using 14 double-differential \ac{EMG} electrodes around the forearm.
(d) Inter- vs.\ intra-session performance varying the number of training sessions $k\in\{2,3,5,7\}$ adopted, showing persistent cross-day domain shift.
(e) Benchmark on NinaPro DB6 comparing accuracy and model size, highlighting the competitiveness of the lightweight \ac{TCN} (0.05M parameters).
}
\label{fig:architecture}
\end{figure*}

Real-world \ac{EMG} applications require robust performance across unseen recording sessions. However, the field often relies on intra-session validation, which artificially inflates performance metrics and obscures the impact of temporal signal drift. This study systematically evaluates intra- and inter-session generalization using a \ac{TCN} decoder~\cite{Zanghieri_etal21} tailored for physiological time-series benchmarked on the NinaPro DB6 dataset~\cite{Palermo_etal17}. 

To reduce the performance loss caused by signal drift, we adapt three complementary test-time adaptation strategies to the \ac{EMG} decoding setting.
First, causal adaptive \ac{BN} provides a lightweight, label-free mechanism that updates normalization statistics online to align test-time inputs with minimal computational overhead. Second, regularized statistical alignment for replay remains unsupervised, but enables adaptation by updating a low-rank subset of parameters while using a replay buffer to preserve representations of the source-domain and limit catastrophic forgetting. This added flexibility comes with higher memory and computational requirements. Third, we employ meta-learning as a supervised few-shot strategy~\cite{Santoro_etal16} that mimics a practical calibration phase in which the user performs a small number of labelled gestures.

Rather than treating these methods as directly comparable alternatives, we use them to characterize the design space of test-time adaptation for \ac{EMG} across a range of deployment constraints, from low-cost unsupervised updates to supervised calibration phases.
Overall, this work connects ideas from \ac{TTT}~\cite{liang2025comprehensive}, \ac{CL}~\cite{parisi2019continual}, and meta-learning~\cite{huisman2021survey}, and adapts these methods to \ac{EMG} decoding. We provide guidance on selecting and deploying adaptation strategies for robust \ac{EMG} decoding, advancing the feasibility of energy-efficient, long-term myoelectric control in wearable applications.

\section{Material and Methods}
\label{sec:m&m}
Fig.~\ref{fig:architecture} shows an overview of the proposed methodology for inter-session adaptation in surface \ac{EMG} decoding.
The study focuses on classifying seven grasping movements from the NinaPro DB6 dataset (Fig.~\ref{fig:architecture}c) using a \ac{TCN} architecture composed of stacked causal convolutional blocks with batch normalization and residual connections (Fig.~\ref{fig:architecture}a–b). The model was trained and evaluated under both intra-session and inter-session conditions to quantify generalization across recording days (Fig.~\ref{fig:architecture}d). In this context, and following previous work, intra-session performance refers to generalisation to unseen samples drawn from the same recording session(s) used for training, whereas inter-session performance refers to generalisation to data collected in a different recording session (e.g. a different day), where signal characteristics may shift due to factors such as electrode placement. In Fig.~\ref{fig:architecture}e, we compare our methodology with previous studies in terms of inter, intra level accuracy and model size. This experimental setup provides a controlled benchmark for assessing adaptation mechanisms that
aim to improve long-term robustness in EMG-based gesture
decoding.

\subsection{Temporal Convolutional Network Architecture}
\label{ssec:tcn}
The proposed decoder is based on a \ac{TCN}, which processes temporal sequences through one-dimensional causal convolutions, effectively capturing correlations between adjacent time samples. 
The proposed decoder is based on a \ac{TCN}, which processes temporal sequences through one-dimensional causal convolutions, effectively capturing correlations between adjacent time samples. \acp{TCN} are particularly well suited to \ac{EMG} signals, as they impose temporal translational equivariance across network operations. They have also recently been shown to achieve state-of-the-art results for \ac{EMG}-based regression~\cite{manneschi2026beyond} in terms of cross-subject generalisation, performance, and computational efficiency.

As shown in Fig.~\ref{fig:architecture}(a–b), the network consists of multiple residual blocks, each containing two causal convolutional layers followed by normalization and non-linear activation. The use of dilation allows each subsequent block to cover a wider temporal context, progressively expanding the receptive field while maintaining computational efficiency. We employ batch normalization~\cite{ioffe2015batch} within each network block to stabilize optimization and accelerate convergence. As discussed in Section~\ref{sssec:BN}, we then introduce a causal variant that adaptively updates normalization statistics online for test-time adaptation. The outputs of the convolutional blocks are projected point-wise through a linear readout layer, generating a response for every input data point.
This residual and dilated design preserves the temporal resolution of the input while enabling hierarchical extraction of temporal dependencies. In more detail, the network is composed of six blocks, each with kernel size 4 and dilation $2^{n-1}$ for block $n = 1, \dots, 6$.

As illustrated in Fig.~\ref{fig:architecture}d, the TCN architecture achieves stable intra-session accuracy, confirming its ability to model short-term temporal structure effectively. However, the same figure highlights a substantial drop in inter-session performance due to cross-day variability and electrode repositioning. This discrepancy motivates the adaptation mechanisms introduced later in the work, which aim to preserve the strengths of the \ac{TCN} while improving its robustness across recording sessions.

\subsection{Dataset and Related works}
\label{ssec:dataset}
Experiments were conducted using the NinaPro DB6 dataset~\cite{Palermo_etal17}, created to investigate the consistency of surface \ac{EMG}-based hand grasp recognition across multiple recording sessions. The experimental setup is illustrated in Fig.~\ref{fig:architecture}c.  The dataset comprises data from ten healthy participants who performed seven distinct grasp types. Each grasp was repeated twelve times per session, with two sessions recorded per day over a period of five consecutive days, yielding ten sessions for each subject. \ac{EMG} activity was measured using fourteen Delsys Trigno double-differential wireless electrodes distributed uniformly around the forearm. The upper eight electrodes were positioned close to the radio-humeral joint, and the remaining six were placed farther down the forearm. Signals were sampled at 2\,kHz. Each trial consisted of a four-second grasp followed by a four-second rest period. The acquisition protocol enables systematic analysis of inter-session variability and temporal adaptation in \ac{EMG}-based gesture decoding.

Fig.~\ref{fig:architecture}e highlights the trade-off between accuracy and model size in NinaPro DB6 gesture recognition. Classical methods such as Random Forests~\cite{Palermo_etal17} are extremely lightweight but exhibit large inter-session drops, whereas TempoNet~\cite{zanghieri2019robust} (460k parameters) improves both intra- and inter-session performance with minimal computational cost. Larger deep models, 
including lightweight transformers~\cite{burrello2022bioformers,chen2025waveformer}, and high-capacity architectures like OTIS~\cite{turgut2024towards} and MOMENT~\cite{goswami2024moment}, achieve strong intra-session accuracy but require hundreds of thousands to millions of parameters and remain sensitive to cross-day variability. In contrast, the proposed adaptive system ($\approx 50$k parameters) offers a superior accuracy–complexity balance, matching larger models intra-session while providing improved inter-session robustness at substantially lower cost. 

\subsection{Validation Methodology}
\label{ssec:validation}
The literature on the NinaPro DB6 dataset uses vastly
different pre-processing and post-processing methodologies to measure model performance
and efficiency, complicating direct comparison.
To evaluate each model under a realistic deployment setting, we partition the ten sessions available for each subject in NinaPro DB6 strictly by time. The first five sessions define the source domain ($\mathcal{D}^{src}$). 
Within $\mathcal{D}^{src}$, we use $60\%$ of the data for training and reserve the remaining $40\%$ to evaluate intra-session performance. The remaining five sessions define the target domain ($\mathcal{D}^{tgt}$), which is used to assess inter-session generalisation. Importantly, sessions are split causally by time, so that all target sessions occur strictly after the source sessions. This reflects a realistic deployment scenario in which a model is trained on past recordings and then evaluated on later sessions subject to electrode repositioning and day-to-day variability.

Prior work typically reports classification performance after removing transient periods around gesture onset and offset, and often relies on study-specific pre-processing and post-processing routines. Accordingly, we adopt this standard transient-removal practice. Furthermore,
to emulate the post-processing smoothing typically seen in applied prosthetic control, we apply a simple post-processing step based on majority voting over a sliding window of predictions. In the same evaluation setting, when studying the benefits of test-time adaptation over our baseline model, we report performance only on active gestures, excluding resting states. This yields a stricter protocol, as improvements cannot be attributed to the typically easier rest class.

In Section~\ref{ssec:comparison}, when comparing our results with previous works in more details, we additionally report complementary performance metrics to account for the heterogeneous pre-processing and post-processing pipelines used in the literature and to provide a more complete view of the system behaviour. Details of these metrics are given in Section~\ref{ssec:comparison}.




\subsection{Adaptation Strategies}
\label{ssec:strategies}
We consider session-based \ac{EMG} data, where each session $s$ is a sequence 
$\mathcal{D}_s=\{(x^{(s)}_t,y^{(s)}_t)\}_{t=1}^{T_s}$, with 
$x^{(s)}_t \in \mathbb{R}^d$ and optional labels $y^{(s)}_t$. 
Sessions are divided into a \textit{source} set $\mathcal{D}^{\text{src}}$ for training and a \textit{target} set 
$\mathcal{D}^{\text{tgt}}$ for evaluation.

At test time, the model processes a target session $\mathcal{D}_{\tau}$ in a streaming fashion and, at each time $t$, has access only to the prefix $\mathcal{D}^{\leq t}_{\tau}$ containing all samples observed up to that point. The model adapts using $\mathcal{D}^{\leq t}_{\tau}$ and is then evaluated on the held-out suffix $\mathcal{D}^{> t}_{\tau}$ from the same session, thereby measuring the benefits of adaptation in a transductive setting.

Moreover, to quantify potential costs of adapting to a specific prefix, we additionally evaluate the adapted model on the remaining target sessions $\mathcal{D}_{\tau'}$ with $\tau' \neq \tau$. We refer to this performance as the \emph{cross session impact}, as it captures how updates induced by the session $\tau$ affect performance in subsequent unseen sessions. In general, we expect adaptation to be detrimental for cross-session impact, since the update is driven by statistics and signal characteristics specific to session $\tau$. We notice how this measure is not intended to model a strict operational cost for a sessioned dataset, since in practice a system may reset its adaptation state when a new session begins. Instead, cross-session impact serves as a diagnostic of update transferability, and becomes relevant when resets are not available or when distribution shifts occur more continuously over time.

To define the prefix $\mathcal{D}^{\le t}_\tau$, we simulated a realistic streaming input by generating
sequences of temporal windows sampled from the session $\tau$.
Sequence duration was set relative to the typical length of a gesture repetition. If the target sequence was shorter than two repetitions, we extracted a single window with a random start
time. If the sequence was longer, we formed it by concatenating
multiple windows, enforcing that each window begins at a
rest state to preserve the original rest–gesture alternation in
the dataset. This shuffling process was necessary to generate
multiple sequences containing different gestures and to ensure
evaluation across different possible gesture orders. Moreover,
this closer mimics the unpredictable nature of daily use, where
the collected data may come from an unknown, unconstrained
segment of routine operation. \newline


We study three levels of test-time data availability:

\begin{enumerate}
    \item \textbf{Unsupervised \ac{TTT}:} Only the unlabelled prefix 
    $\mathcal{D}^{\le t,\text{unlab}}_\tau$ is available, and no source data can be accessed at test-time.

    \item \textbf{Resource-aware adaptation:} The model additionally uses a small exemplar buffer 
    $\mathcal{D}_{\text{er}} \subset \mathcal{D}^{\text{src}}$, having access to a constrained number of source data.

    \item \textbf{Few-shot learning:} The unlabelled prefix is augmented with a sparse set of labels 
    $\mathcal{D}^{\le t,\text{lab}}_\tau$, simulating minimal user-assisted calibration.
\end{enumerate}

Relevant to settings (1)--(2), a key distinction from standard \ac{TTT} benchmarks is that the target \ac{EMG} arrives
as a continuous and highly noisy stream, with limited data available for adaptation. In this regime, adaptation cannot
assume consistently reliable gesture segmentation, making it difficult to form semantically coherent mini-batches or to define label-preserving (or, in other words, physiologically plausible) augmentations. Consequently, contrastive self-supervised objectives (e.g., SimCLR-style learning)
are often ill-suited: naïve positive/negative pair construction may mix heterogeneous activations, and common augmentations
can distort physiologically meaningful structure. Moreover, individual \ac{EMG} samples are weakly informative in isolation
and require temporal context for stable updates. For these reasons, we focus on adaptation via statistical alignment
between source and target feature distributions, which remains well defined under streaming, noisy, and low-data constraints.
For few-shot learning (setting (3)) under supervised calibration phases, we instead rely on meta-learning, which explicitly
optimizes the model for rapid adaptation from a small number of labelled examples. We finally note that the performance obtained through these different methodologies is not directly comparable, given their different assumptions about data availability. For instance, the few-shot learning regime assumes the availability of labels, and it is therefore expected to achieve better performance than the other methods. However, the rationale for exploring these different settings is to show that test-time adaptation, as a general principle, can be effectively applied to EMG decoding, and how, depending on the specific scenario, energy constraints and data availability, some methods are more suitable than others.\\

\subsubsection{\textbf{Adaptive Batch Normalization}}
\label{sssec:BN}
This section addresses the first regime in Section~\ref{ssec:strategies}, i.e., \ac{TTT}, where only unlabelled target data are available. To enable adaptation under this constraint, we introduce an online, causal variant of adaptive \ac{BN}. The model aligns at test time by updating only its internal normalization statistics (mean and variance), gradually shifting them from the source to the target domain while leaving the parameters unchanged.
Classical \ac{BN} computes per-channel statistics over a mini-batch of $B$ sequences of length $U$. Denoting 
$z^{(l)}_{b,c,u}(\tau)$ as the activation at layer $l$, channel $c$, sample $b$, and time $u$ in batch $\tau$, the batch 
mean, and variance are
\begin{align}
\mu^{\text{batch}}_{c}(\tau)=\frac{1}{BU}\sum_{b=1}^{B}\sum_{u=1}^{U} z^{(l)}_{b,c,u}(\tau), \\
(\sigma^{\text{batch}}_{c}(\tau))^2=\frac{1}{BU}\sum_{b=1}^{B}\sum_{u=1}^{U}(z^{(l)}_{b,c,u}(\tau)-\mu^{\text{batch}}_{c}(\tau))^2.
\end{align}
These are used to normalize activations:
\begin{equation}
\hat z^{(l)}_{b,c,u}=\frac{z^{(l)}_{b,c,u}-\mu^{\text{batch}}_{c}}{\sqrt{(\sigma^{\text{batch}}_{c})^2+\varepsilon}},\qquad
y^{(l)}_{b,c,u}=\gamma^{(l)}_{c}\hat z^{(l)}_{b,c,u}+\beta^{(l)}_{c},
\label{eq:bn_apply}
\end{equation}
while long-term population statistics are tracked via \acp{EMA}.

In the streaming test-time setting, we process a single sequence ($B=1$) and compute online statistics over the prefix 
$\{z_c(i)\}_{i=1}^{t}$. At time $t$:
\begin{align}
\mu_c(t) = \frac{1}{t}\sum_{i=1}^{t} z_c(i), \qquad
\sigma_c^2(t) = \frac{1}{t}\sum_{i=1}^{t}(z_c(i)-\mu_c(t))^2,
\end{align}
which can be updated incrementally using Welford's method:
\begin{align}
\mu_c(t)=\mu_c(t-1)+\frac{z_c(t)-\mu_c(t-1)}{t}, \\
S_c(t)=S_c(t-1)+(z_c(t)-\mu_c(t-1))(z_c(t)-\mu_c(t)), \\
\sigma_c^2(t)=\frac{S_c(t)}{t}.
\end{align}

To retain information from the source domain, we blend the source \ac{EMA} statistics with the online test-time
estimates using a mixing coefficient $\alpha\in[0,1]$:
\begin{align}
\mu_c^{\text{use}}(t) &= (1-\alpha)\,\mu_c^{\text{EMA}} + \alpha\,\mu_c(t), \\
(\sigma_c^{\text{use}}(t))^2 &= (1-\alpha)\,(\sigma_c^{\text{EMA}})^2 + \alpha\,\sigma_c^2(t).
\end{align}

In the fully online setting, $\alpha$ should reflect the reliability of the target estimates, which increases as more
samples are observed. We therefore adopt an adaptive, causal schedule
\begin{equation}
\alpha(t)=\min\!\left(1,\;\frac{\beta}{N}\,t\right),
\label{eq:alpha_schedule}
\end{equation}
where $t$ is the number of target samples seen so far, $N$ denotes the (expected maximum) number of samples in the target
session, and $\beta$ is a scalar controlling the overall adaptation strength. This schedule gradually shifts the
normalization from the source \ac{EMA} statistics towards the target online estimates. 
In our analysis, we study the effect of this adaptive causal \ac{BN} by explicitly controlling $t$ and $\alpha$
(either fixing $\alpha$ or using Eq.~\eqref{eq:alpha_schedule} to tie it to the observed prefix length). During
deployment, the model updates $\alpha(t)$ online as new target samples arrive and uses the blended statistics
$\{\mu_c^{\text{use}}(t), \sigma_c^{\text{use}}(t)\}$ in place of the static \ac{EMA} statistics in
Eq.~\eqref{eq:bn_apply}.\\

\subsubsection{\textbf{Statistical Alignment}}
\label{sssec:stat_alig}

This section corresponds to the second regime in Section~\ref{ssec:strategies}, where a small exemplar buffer
$\mathcal{D}_{\text{er}}$ is available at test-time. We perform resource-efficient online adaptation by updating only
low-rank adaptation parameters through LoRA \cite{hu2021loralowrankadaptationlarge}, while keeping the backbone weights frozen. Adaptation is driven
by statistical alignment in latent feature space, and the exemplar buffer is used as a regularizer to stabilize updates
and mitigate forgetting.

Let $z^{(l)}\in\mathbb{R}^{d_l}$ denote the latent representation at layer $l$. At test-time, we maintain an online
buffer $Z_\tau=\{z_i\}_{i=1}^{M}$ of recent target features (extracted with the current model), from which the target
statistics required by the chosen alignment objective are computed. We consider three alignment objectives: 

\textit{Global moment matching (\textsc{Cov}).}
We summarize the source feature distribution at layer $l$ with its mean and covariance,
\begin{align}
\mu_{\text{src}} = \mathbb{E}_{\mathcal{D}^{\text{src}}}[z], \qquad
\Sigma_{\text{src}} = \mathbb{E}_{\mathcal{D}^{\text{src}}}\!\left[(z-\mu_{\text{src}})(z-\mu_{\text{src}})^{T}\right],
\end{align}
and compute empirical target moments $(\mu_\tau,\Sigma_\tau)$ from $Z_\tau$. Adaptation is driven by
\begin{align}
\mathcal{L}_{\text{align}}^{\text{mom}}
= \lambda_{\mu}\|\mu_{\tau} - \mu_{\text{src}}\|_2^2
+ \lambda_{\Sigma}\|\Sigma_{\tau} - \Sigma_{\text{src}}\|_F^2.
\end{align}
In the following, we refer to this variant as \textsc{Cov}, reflecting the covariance alignment it performs.

\textit{Class-conditional moment matching.}
We partition the source features into rest ($R$) and gesture ($G$) states, and estimate class-specific moments
$(\mu_{\text{src},k},\Sigma_{\text{src},k})$ for $k\in\{R,G\}$. At test-time, pseudo-labels assign each feature in
$Z_\tau$ to $k\in\{R,G\}$, producing subsets $Z_{\tau,k}$ and corresponding empirical moments
$(\mu_{\tau,k},\Sigma_{\tau,k})$. We then align moments per state:
\begin{align}
\mathcal{L}_{\text{align}}^{\text{cc}}
= \sum_{k\in\{R,G\}}
\lambda_{\mu}\|\mu_{\tau,k} - \mu_{\text{src},k}\|_2^2
+ \lambda_{\Sigma}\|\Sigma_{\tau,k} - \Sigma_{\text{src},k}\|_F^2.
\end{align}

\textit{Distribution matching via \ac{GMM}.}
Instead of restricting to low-order moments, we fit a \ac{GMM} $\mathcal{G}_{\text{src}}$ to source features at layer
$l$. During adaptation, we draw samples $\hat{Z}_{\mathcal{G}_{\text{src}}}$ from $\mathcal{G}_{\text{src}}$ and match
their distribution to the target buffer $Z_\tau$ using the \ac{SWD}:
\begin{align}
\mathcal{L}_{\text{align}}^{\text{gmm}} = \text{SWD}\!\left(\hat{Z}_{\mathcal{G}_{\text{src}}},\, Z_\tau\right).
\end{align}

To mitigate catastrophic forgetting while updating $\theta_{\text{LoRA}}$, we regularize adaptation using exemplars
$(x_{\text{er}},y_{\text{er}})\in\mathcal{D}_{\text{er}}$ following \ac{DER}. For each exemplar, we store the source model
output $\hat{y}_{\text{er}}$ (e.g., logits) and penalize deviations under the adapted model, optionally combining this
with supervised replay:
\begin{align}
\mathcal{L}_{\text{er}}
= \|\hat{y}(x_{\text{er}})-\hat{y}_{\text{er}}\|_2^2
\;+\;
\mathcal{L}_{\text{CE}}(\hat{p}(x_{\text{er}}), y_{\text{er}}),
\end{align}
where $\hat{y}(\cdot)$ denotes current model outputs, $\hat{p}(\cdot)$ the corresponding class probabilities, and
$\mathcal{L}_{\text{CE}}$ is the cross-entropy loss.

To control the magnitude of test-time updates induced by statistical alignment, we restrict adaptation to \ac{LoRA}
parameters inserted in the convolutional layers. Denoting by $\phi$ the collection of trainable \ac{LoRA} weights, we
optimize 
\begin{align}
\mathcal{L}(\phi)
= \alpha\,\mathcal{L}_{\text{align}} + \beta\,\mathcal{L}_{\text{er}},
\end{align}
where $\mathcal{L}_{\text{align}}\in\{\mathcal{L}_{\text{align}}^{\text{mom}},\mathcal{L}_{\text{align}}^{\text{cc}},\mathcal{L}_{\text{align}}^{\text{gmm}}\}$ specifies the chosen alignment objective.\\

\subsubsection{\textbf{Meta-learning}}
\label{sssec:meta_learning}
This section corresponds to the third regime in Section~\ref{ssec:strategies}, where a small number of supervisory
signals is available at test-time, i.e. few-shot settings. Similarly to the statistical alignment methods of the previous paragraph, adaptation will be performed over the LoRA parameters. This regime quantifies the performance gains achievable
through a short calibration phase, during which a user could repeat a small set of gestures to personalize the decoder. 

To exploit limited supervision efficiently, we adopt gradient-based meta-learning in the model-agnostic meta-learning
(MAML) framework, tailored to sessioned \ac{EMG} data. We treat each session as a task $\mathcal{T}_s$ and learn a network initialization that can be adapted to a new session using only a few labelled calibration examples.

We parameterize the model as
$f(x;\theta,\phi)$, where $\theta$ are \emph{frozen} backbone parameters and $\phi$ are the \ac{LoRA} parameters. Task-specific adaptation operates only on $\phi$, while $\theta$ remains fixed. Therefore, the meta-learned initialization consists of a shared \ac{LoRA} starting point $\phi$ (together with the fixed
$\theta$ learned on $\mathcal{D}^{\text{src}}$).

For each task/session $s$, we split labelled samples into a support (calibration) set $\mathcal{D}^{\text{sup}}_s$ and a
query set $\mathcal{D}^{\text{qry}}_s$. Let $\mathcal{L}(\theta,\phi;\mathcal{D})$ denote the supervised loss (cross
entropy). Starting from the meta-initialization $\phi$, the inner-loop adaptation performs $K$ gradient steps on the support set.
For notational simplicity, we omit the explicit dependence of $\mathcal{L}$ on the support data
$\mathcal{D}^{\text{sup}}_s$ (i.e., $\mathcal{L}(\theta,\phi;\mathcal{D}^{\text{sup}}_s)$):
\begin{align}
\phi^{(0)}_s &= \phi, \\
\phi^{(k)}_s &= \phi^{(k-1)}_s - \eta_i \nabla_{\phi^{(k-1)}_s}
\mathcal{L}\!\left(\theta,\phi^{(k-1)}_s \right),
\ \ \ k=1,\dots,K,
\label{eq:maml_lora_inner}
\end{align}
where $\eta_i$ is the inner-loop learning rate.

The outer-loop meta-objective then updates the shared initialization $(\theta,\phi)$ such that the adapted parameters
$\phi'_s \equiv \phi^{(K)}_s$ (obtained via Eq.~\eqref{eq:maml_lora_inner}) generalize on the corresponding query sets:
\begin{align}
\min_{\theta,\phi}\;\; \sum_{\mathcal{D}_s \in \mathcal{D}^{\text{src}}}
\mathcal{L}\!\left(\theta,\phi'_s\right),
\label{eq:maml_lora_obj}
\end{align}
Using a meta-learning rate $\eta_o$, we perform the
meta-update.
\begin{align}
(\theta,\phi) \leftarrow (\theta,\phi)
-\eta_o \nabla_{(\theta,\phi)}
\sum_{\mathcal{D}_s \in \mathcal{D}^{\text{src}}}
\mathcal{L}\!\left(\theta,\phi'_s;\mathcal{D}^{\text{qry}}_s\right).
\label{eq:maml_lora_outer}
\end{align}

The difference between first- and second-order meta-learning lies in how the meta-gradient in
Eq.~\eqref{eq:maml_lora_outer} is computed. In second-order MAML, the gradient
$\nabla_{(\theta,\phi)} \mathcal{L}(\theta,\phi'_s;\mathcal{D}^{\text{qry}}_s)$ backpropagates through the $K$ inner-loop
updates in Eq.~\eqref{eq:maml_lora_inner}, and therefore includes second-order terms arising from differentiating through
the inner-loop gradients (i.e., Hessian--vector products). In first-order MAML, these higher-order
terms are neglected by treating $\phi'_s$ as a constant with respect to $(\theta,\phi)$ when computing the meta-update.
Equivalently, second-order MAML can be viewed as learning an initialization by optimizing through the unrolled
$K$-step adaptation dynamics, analogous to backpropagation through time in a recurrent system.

To mitigate forgetting arising from rapid adaptation, we also evaluate a second variant that combines meta-learning with replay-based regularization. Specifically, we incorporate a \ac{DER}-style term using a small exemplar memory analogously to the replay-regularized statistical alignment methods.

\section{Results}
\label{sec:results}

\subsection{Baseline results}
We first evaluated the baseline performance of the proposed \ac{TCN} model on the NinaPro DB6 dataset to assess its ability to generalize across recording sessions. The network was trained to classify seven grasping gestures from multi-day \ac{EMG} recordings, with ten sessions available for each subject. Two evaluation schemes were considered: intra-session, where training and testing data belong to the same session, and inter-session, where testing is performed on unseen sessions to evaluate cross-day generalization.

Fig.~\ref{fig:architecture}d reports the error rates ($1-$accuracy) for both intra-session and inter-session evaluations as a function of the training iterations and the number of training sessions $k$ (shown by different colored lines), where we vary $k \in \{2,3,5,7\}$. 
With this protocol, the intra-session training and testing samples are drawn from the same sessions used for optimisation, and accuracy remains above $79\%$ across all values of $k$, ranging from $85.6\%$ when seven training sessions are used to $79.3\%$ when only two are available. This indicates that the network reliably captures temporal patterns under stationary conditions, even when trained across as few as two sessions. 
In the inter-session case, where the model is evaluated on entirely new sessions, accuracy drops to $56.7\%$ even in the most favourable case of seven training sessions, reflecting the substantial domain shift caused by electrode repositioning and day-to-day physiological variability. \color{black} As more sessions are included in training, inter-session performance progressively improves, indicating that exposure to multiple recording conditions enhances generalisation to unseen data. However, even on the Ninapro DB6 dataset, which was explicitly collected to study inter-session variability and is, to our knowledge, the largest open-source \ac{EMG} dataset with repeated sessions, the drop in performance from intra-session to inter-session evaluation exceeds $20\%$, validating the necessity to focus on inter-session performance. 

\subsection{Adaptive Batch Normalization}
\label{ssec:adaptbatcnor}

\begin{figure*}[ht!]
\centering
\includegraphics[width=1.\textwidth]{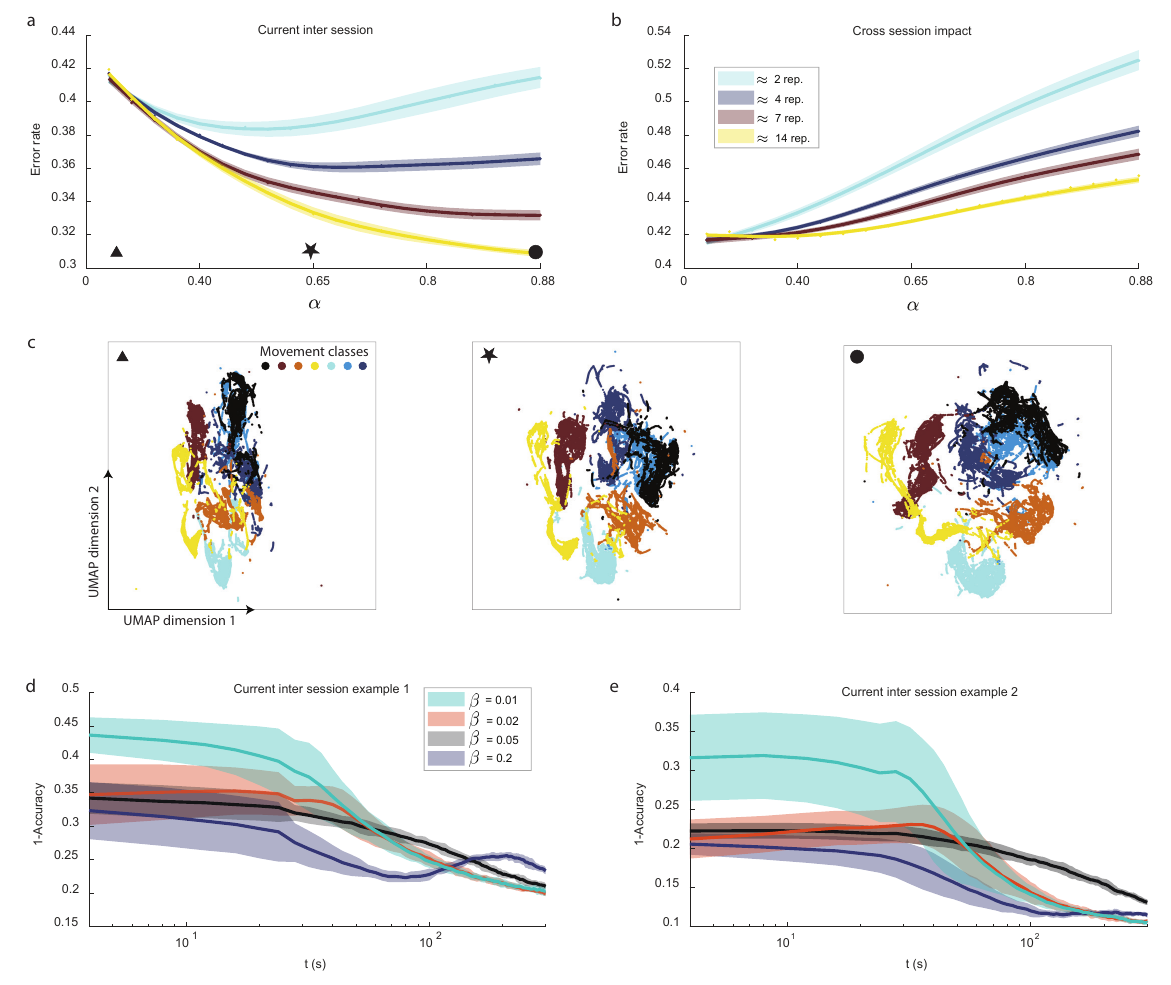}
\caption{Performance of the causal variant of batch normalization.
(a,b) Batch-normalization adaptation: evolution of error rate on the current session (a) and on all other sessions (b) as a function of the mixing coefficient between old and new statistics. Results are reported for 2, 4, 7, and 14 gesture repetitions used for adaptation. (c) illustrates the shift of the UMAP projections across the adaptation strengths. (d,e) Online adaptation performance: evolution of the error rate over time for two representative subjects with different colors representing different adaptation speeds $\beta$.}
\label{fig:adaptation_results}
\end{figure*}

We next evaluated the effectiveness of adaptive batch normalization (Fig.~\ref{fig:adaptation_results}). Each session was treated as a target domain unseen during training, and adaptation was performed by blending the target and source batch-normalization moments as described in
Section~\ref{sssec:BN}.

Figs.~\ref{fig:adaptation_results}a,b report performance as a function of the blending coefficient $\alpha$ for four adaptation-prefix sizes corresponding to 2, 4, 7, and 14 gesture repetitions (spanning approximately 8--112\,s of data). For each target session, repetitions are sampled to form an unbalanced sequence of gestures, as described in Section~\ref{ssec:strategies}, to mimic a realistic test-time data stream. After adapting on this prefix, we evaluate performance either on the remaining samples from the same session, quantifying the transductive benefit of adaptation, or on samples from different sessions, quantifying cross-session impact (Section~\ref{ssec:strategies}).

 The left panel shows the evolution of the error rate on the current session as the model progressively replaces the source \ac{BN} statistics with target estimates. This is controlled by the parameter $\alpha$, which weights the contribution of newly observed moments when updating the stored mean and variance learned on the training data. When only limited adaptation data are available (two or four repetitions; light and dark blue curves in Fig.~\ref{fig:adaptation_results}a), overly large values of $\alpha$ can degrade performance, because early moment estimates poorly represent the full gesture distribution. With larger adaptation sets (7 or 14 repetitions), the target statistics become more representative, allowing the model to recover to approximately $69\%$ accuracy.

Overall, performance depends jointly on $\alpha$ and the amount of test-time data, revealing a trade-off in which smaller $\alpha$ values are preferable when only a few repetitions are observed, whereas larger $\alpha$ values become beneficial once more adaptation data are available. These results motivate an adaptive batch-normalisation variant in which $\alpha$ is selected as a function of the amount of evidence accumulated at test time. 

The right panel reports cross-session impact in terms of error rate. It shows an increase in error on held-out sessions as the model specialises to the current session, which is an expected side effect of session-specific adaptation.

Figs.~\ref{fig:adaptation_results}c,d illustrate how class structure in the latent space evolves as adaptation progresses. We report the case with 14 repetitions and highlight three increasing adaptation strengths (triangle, star, and circle, respectively). Latent features are visualized with Uniform Manifold Approximation and Projection (UMAP) \cite{mcinnes2018umap}, showing progressively tighter clusters and improved class separability as adaptation strengthens.

To enable a consistent comparison across time steps and adaptation strengths, we compute a shared UMAP embedding and use it to project features extracted at different adaptation points (rather than fitting UMAP independently for each plot). We also regularize the visualization by discouraging large shifts of the latent representations across time, so that changes reflect genuine improvements in separability rather than arbitrary embedding drift. Finally, we note that perfect separation is not expected: the \ac{EMG} stream transitions continuously between gestures, and boundary segments naturally exhibit overlapping activations.

Finally, Figs.~\ref{fig:adaptation_results}d,e show online batch-normalization adaptation, where the blending coefficient $\alpha$ is now a temporally dependent variable $\alpha(t)$ based on Eq.~\ref{eq:alpha_schedule}. Under our causal test-time formulation, we can track the error rate as the model adapts continuously and $\alpha(t)$ increases with the amount of target data observed. Results are shown for two representative subjects; for each subject, we average the online error over $\sim$100 test-time runs. This online setting is equivalent to a continuously growing prefix $\mathcal{D}^{\leq t}_{\tau}$, where at each step the model predicts the next sample and then updates its \ac{BN} statistics using the newly observed data.

As expected, $\beta$ controls the adaptation timescale: larger $\beta$ produces faster but less stable updates, whereas smaller $\beta$ produces slower but smoother improvements. Consistent with this interpretation, the most aggressive setting (dark blue) exhibits noticeable oscillations in panel~d, indicating that short-term statistics are noisy and can temporarily over-correct the normalization before converging as more target data accumulate. We note that the curves start from different error rates because some adaptation has already occurred while the TCN receptive field (RF) is being filled. Consequently, by the time the first prediction is made (once the RF is full), each variant has already updated the running statistics to a different extent according to its $\beta$ value, leading to different initial performance.

\subsection{Statistical Alignment}
\label{ssec:statisticalAlignment}

\begin{figure*}[ht!]
\centering
\includegraphics[width=1.\textwidth]{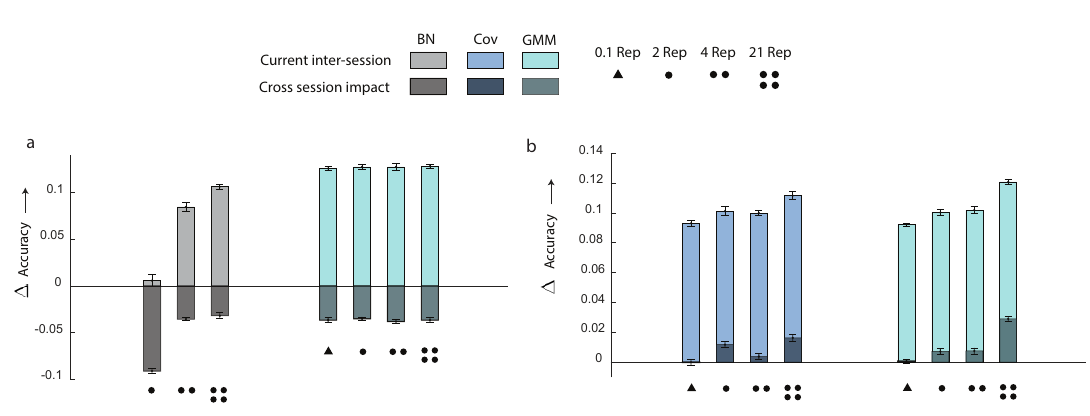}
\caption{Performance of the statistical alignment strategy across sessions. 
(a) shows the performance of the batch-normalization adaptation and GMM adaptation without a DER component, while (b) compares the performance of the Covariance and GMM approaches with a DER component. The bar plots are achieved after convergence of the adaptation process. }
\label{fig:adaptation_stats_results}
\end{figure*}

We now turn to statistical alignment methods, analyzing how adaptation granularity, from matching low-order moments in feature space to fitting a \ac{GMM} to the distribution, affects performance, and how a replay buffer mitigates forgetting. During preliminary exploration, we experimented with aligning statistics by minimising a cost function (defined in Section~\ref{sssec:stat_alig}) at the representation produced by the fourth block of the architecture, as well as by minimising the average of the cost functions defined across the first four blocks. We found that aligning a single cost function is sufficient to achieve the best overall performance. A plausible explanation is that, once the later representation is aligned, upstream feature distributions are implicitly regularised through the shared forward pass, making additional per-block losses redundant. The minimisation is performed over rank-4 LoRA parameters in the blocks preceding the chosen alignment point.


In this case, we focus on the final performance achieved after the adaptation process, which we observe to converge in approximately $100$ gradient steps. Adaptation is performed by aligning the responses computed on the prefix $\mathcal{D}_{\tau}^{\leq t}$ to the training-data statistics.

Figs.~\ref{fig:adaptation_stats_results}a,b summarise the resulting performance gains over the baseline using bar plots. We report both inter-session performance (lighter colours) and cross-session impact (darker colours) as a function of the prefix size used for alignment, measured in gesture repetitions. Panel~a compares the GMM-based alignment against batch-normalisation alignment, while panel~b contrasts covariance matching (Cov) and the GMM variant when an experience-replay buffer is used as a regulariser.

Compared to adaptive batch normalisation (Fig.~\ref{fig:adaptation_stats_results}a), statistical alignment is less sensitive to small prefixes $\mathcal{D}_{\tau}^{\leq t}$. Batch normalisation yields negligible improvements over the baseline when only $0.1$ gesture repetitions are used for adaptation ($0.1$~Rep), and it instead produces a cross-session impact that drops by $0.09$ relative to the baseline. Moreover, as the prefix size increases from $0.1$ to $21$ repetitions, the inter-session accuracy of batch normalisation varies substantially, reaching a $0.12$ improvement at $21$ repetitions. In contrast, GMM alignment remains surprisingly stable, providing an accuracy gain of approximately $0.13$ for both $0.1$ and $21$ repetitions. This indicates that matching richer feature statistics provides a more informative alignment signal than mean- and variance-based \ac{BN} updates, making the method more robust when only limited adaptation data are available. Nevertheless, cross-session impact remains negative in panel~a, consistent with adaptation that specialises to the current streaming session.

We now turn to Fig.~\ref{fig:adaptation_stats_results}b, where we include an experience-replay regulariser. In contrast to panel~a, replay helps preserve source-domain knowledge, and the resulting cross-session impacts are no longer negative. Instead, as the amount of data used for alignment increases, cross-session impact improves for both covariance matching (blue bars) and the GMM-based method (green bars). Inter-session performance also increases more steadily than in the non-regularised setting as additional gesture repetitions are used.

A natural interpretation is that larger prefixes yield more reliable gradient estimates, whose update directions are more compatible with previously learned representations. As a result, adaptation becomes increasingly beneficial as more target data are available. The improved cross-session impact further indicates that the statistical-alignment updates are not purely session-specific. Rather, aligning latent statistics while replaying source exemplars appears to move the model toward feature representations that are robust to session-to-session variability, instead of fitting idiosyncrasies of the current target stream. More broadly, these results highlight the complementarity between test-time self-supervision (via unlabelled statistical alignment) and continual-learning mechanisms (via replay): together, they incorporate useful information from the target stream while maintaining, and in some cases improving, generalisation.

Overall, the \ac{GMM} approach achieves higher accuracy than covariance matching (Fig.~\ref{fig:adaptation_stats_results}b), consistent with its ability to capture finer-grained structure in the feature distribution beyond low-order statistics. The regularised GMM yields inter-session and cross-session gains of $0.09$ and $0.01$ at $0.1$ repetitions, and $0.12$ and $0.03$ at $21$ repetitions, respectively.

\subsection{Meta-learning}

Finally, we turn to meta-learning. The goal is to quantify the gains achievable through fast \emph{supervised} adaptation, thereby mimicking a short user calibration phase. Since labels provide a stronger adaptation signal, we expect supervised few-shot adaptation to provide larger performance gains than self-supervised strategies.
 At the same time, considering few-shot regimes raises a practical question: can the system recover reliably under OOD shifts with \emph{less than one shot} per gesture, that is, when the available calibration data are sparser than a single labelled example for every class?

We perform these experiments using the MAML algorithm~\cite{finn2017model}, with four inner-loop gradient steps during meta-training. As in Section~\ref{ssec:statisticalAlignment}, test-time adaptation (here, the inner-loop updates) is applied only to rank-4 LoRA parameters.

\begin{figure*}[ht!]
\centering
\includegraphics[width=0.95\textwidth]{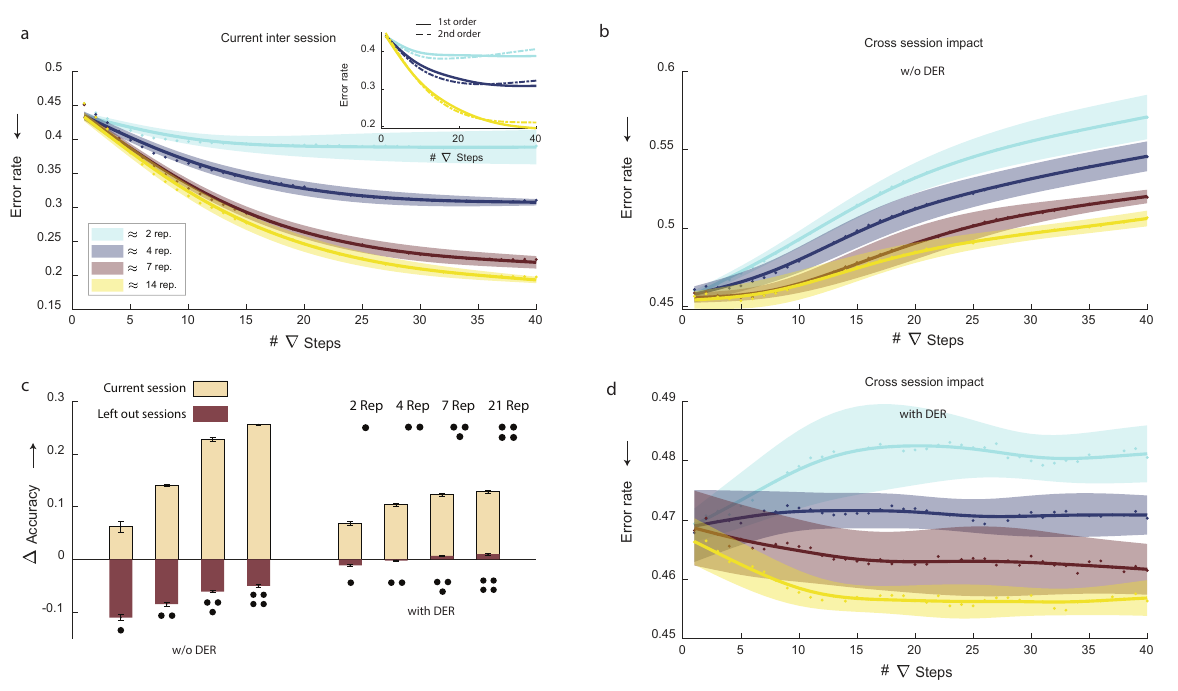}
\caption{Performance of meta-learning across sessions. (a,b) illustrate the evolution of the error rate of the current session (a) and other sessions (b) across adaptation steps (x-axis) without DER and with respect to buffer size. The same data points are used for the different steps of the adaptation, as in few-shot learning settings. A comparative bar plot is made in (c) to display the effect of buffer size and DER on the final accuracy. Finally, (d) showcases the left-out session error rate when using the DER.}
\label{fig:meta}
\end{figure*}

Figure~\ref{fig:meta}a,b report adaptation trends analogous to Section~\ref{ssec:adaptbatcnor}, showing inter-session accuracy and cross-session impact during test-time adaptation. As expected, access to labels yields rapid convergence, typically within $30$--$40$ gradient steps. Moreover, accuracy on the adapted session (Fig.~\ref{fig:meta}a) surpasses $0.8$ when more than one example per gesture is available. 

At the same time, the $\sim 10\%$ performance gap between the 4- and 7-repetition setting highlights the importance of having at least one labelled example per class for reliable calibration (one-shot corresponds to 7 repetitions).

The inset in Fig.~\ref{fig:meta}a compares first-order (solid) and second-order (dashed) MAML variants, where second-order MAML typically converges faster (in approximately $20$ gradient steps). This speedup of second-order MAML, however, can come with increased sensitivity to the number of adaptation steps, consistent with a higher tendency to overfit, an effect that can be attributed to the fact that second-order gradients effectively shape the update dynamics, making the optimization process itself part of what is learned. In addition, because supervised calibration explicitly optimizes performance on the target session, meta-learning can become more session-specific, which manifests as a performance drop in terms of cross-session impact (Fig.~\ref{fig:meta}b). As in prior experiments, increasing the number of calibration repetitions mitigates this effect, yielding less over-specialization.

To further reduce the performance loss in cross-session impact, we also augment meta-learning with an experience replay buffer. This stabilizes the calibration updates and preserves performance on sessions not used for adaptation (Fig.~\ref{fig:meta}d), albeit typically at the cost of lower accuracy on the adapted session. Figure~\ref{fig:meta}c summarizes these results: left bars correspond to standard meta-learning, while right bars report the replay-regularized variant. 
Finally, to assess the benefit of framing the problem as meta-learning, we also implement a fine-tuning baseline. In this case, the model is not meta-trained; instead, we adapt a conventionally pre-trained baseline using the same few-shot calibration data. The results are reported in Table~\ref{tab:all_results}, where MAML achieves higher accuracy than this simpler fine-tuning approach.

\subsection{Comparison with State-of-the-art and Compute/Memory Trade-offs}
\label{ssec:comparison}

Table~\ref{tab:all_results} compares our models with prior work, focusing on inter-session accuracy. For consistency with
the literature, we report performance on the \emph{current} adaptation session and omit variants that use the \ac{DER}
buffer, which is intended to improve the system in terms of retention and cross-session impact. To enable comparison with the literature and provide a more detailed view of system behaviour, we report: (i) accuracy after transient removal, (ii) accuracy computed on active movements while excluding the rest state after transients removal (the setting used to assess the benefits of adaptation in the previous sections), and (iii) overall accuracy without any pre- or post-processing, on the overall dataset.

Despite its simplicity, the base \ac{TCN} already improves upon earlier temporal baselines, and unlabelled test-time
adaptation yields further gains. \ac{AdaBN} achieves strong improvements with negligible additional memory and MMAC
overhead, since it only updates normalization moments. The \ac{GMM}-based alignment also improves performance while
remaining relatively lightweight: in our accounting, it requires only $\sim$68\,kB of additional memory and $\sim$1.2\,MMAC extra
compute beyond the forward pass. These compute figures are reported for an example batch size of 32 and a sequence length
of 1 to match common reporting in the literature. MAML can further improve accuracy, but it requires a longer prefix of labelled data and higher computational resources. Nevertheless, meta-learning offers a clear advantage over simpler supervised baselines that do not include a meta-training stage (Table~\ref{tab:all_results}).

We note, however, that this unit-length accounting does not correspond to a full adaptation update in practice. In statistical alignment with \ac{GMM} or moments matching (Cov), the model typically requires multiple gradient steps (approximately $100$) on the alignment objective before convergence.
Consequently, a complete adaptation step over an entire gesture incurs a proportionally larger cost
(e.g., $\sim$10.1\,MMAC), but this increase is comparable to processing the same time horizon with a standard forward
pass. Finally, the compute gap between \ac{GMM}-based alignment and meta-learning stems from where adaptation is applied:
statistical alignment updates through minimisation of a cost function defined at intermediate representation (at the fourth
block), whereas meta-learning requires backpropagation through the full network during the inner-loop gradient steps.

\begin{table*}[th!]
 \centering
    \caption{\small{Performance and Resource Usage Comparison with State-Of-the-Art Methods \newline
    MMAC are reported per time step (sequence length 1) with an example batch size 32, separating the base model and adaptation cost (in parenthesis).
    \newline Adaptation mechanisms were used on the out-of-distribution inter-session data.
    }}
    \label{tab:all_results}

\renewcommand{\arraystretch}{1.3}
 \begin{tabular}{| >{\centering\arraybackslash} m{1.4in} *6{| >{\centering\arraybackslash} m{0.75in}} |}
 \hline
\textbf{Model} & Accuracy w/o Transients\newline Inter (Intra) & Accuracy \newline w/o Rest States and Transients\newline Inter (Intra)  &  Full Session Accuracy \newline Inter (Intra) & Parameters & Memory &  MMAC \\ [0.5ex]
 \hline
 \textbf{Previous Works}& & & & & &\\

TempoNet \cite{zanghieri2021semg} & $65.2$ \ (71.8) & ---& 49.6 \ (54.5) & $460$k & 1.8\,MB & $16$ \\
ECNN-A \cite{lin2023robust} & --- & --- & $51.4$ \ \ (---) & ---& --- & --- \\
Bioformer  \cite{burrello2022bioformers}  & $65.7$ \ \ (---) & --- &---& $6.5$k & 94.2 kB & $5.3$ \\
Waveformer \cite{chen2025waveformer} & --- \ \  ($81.9$) & --- & --- & 3.10M & --- & ---\\
 \hline
 \textbf{This work} & & & & & &\\
Base Model, TCN & 74.3 \ (87.7) & 56.6 \ (85.1) & 59.4 \ (78.6) & 47k & 200\,kB & 1.5 \\ 
\hline
 \textbf{This work} \ \ \ \ \ \ \  (Unsupervised adaptation) & & && & &\\
 Adaptive BN & 82.0 & 68.9 & 67.9 & 47k & 200\,kB & 1.5   \\ 
 GMM  & 81.9 & 69.8  & 67.6 &48.7k & 268.2\,kB (4 Gestures) & 1.5 + (1.2)  \\ 
\hline
\textbf{This work}  \ \ \ \ \ \ \  \ \ \ \ (Supervised adaptation) & & & & & &\\
 Fine-tuning  & 84.1  & 77.8 & 70.0 &48.7k & 558.4\,kB (21 Gestures) & 1.5 + (1.5) \\
 MAML  & 87.0   & 80.3 & 74.1 &48.7k & 558.4\,kB (21 Gestures) & 1.5 + (1.5) \\ 
 \hline
\end{tabular}
\end{table*}

\section{Discussions}
\label{sec:discussions}

Achieving stable and generalizable decoding of \ac{EMG} signals across sessions is essential for the reliable use of myoelectric interfaces in real-world, long-term applications. Our experiments highlight distinct adaptation behaviors across the proposed strategies in inter-session \ac{EMG} decoding. The task consists of classifying seven grasping movements from the NinaPro DB6 dataset, which is challenging due to the similarity between gestures and the strong temporal variability of \ac{EMG}. Without adaptation, the baseline \ac{TCN} achieves roughly 85\% intra-session accuracy but drops to 56.6\% inter-session, confirming the limited generalization of \ac{EMG}-based decoders.

The first method, based on causal test-time updates of batch-normalization statistics, provides a lightweight, unsupervised mechanism to recalibrate feature distributions without retraining, making it well suited for edge or wearable deployment. Its performance, however, depends strongly on the amount of available adaptation data: with only a few repetitions, target normalization statistics are poorly estimated, which can cause a transient accuracy drop below the baseline and mild over-specialization to the current session. Because updates are performed online, we can track performance continuously as target statistics accumulate; in practice, reliable recovery typically requires a sufficiently long horizon at test-time (on the order of tens of seconds) to obtain stable estimates. This approach remains computationally inexpensive, as \ac{BN} incurs negligible additional MAC operations relative to the full network. Evaluating performance on other test sessions while adapting to the current one (cross session impact) nevertheless reveals a small deterioration, consistent with the model becoming increasingly session-specific. From this perspective, \ac{BN} adaptation can be viewed as a form of statistical alignment, where features are re-centered and re-scaled (towards zero mean and unit variance) using target-domain estimates.

Turning to explicit statistical alignment with a tiny experience replay buffer in the spirit of dark experience replay (\ac{DER}), we observe a complementary behavior: adaptation to the current session remains effective while performance on sessions not used for adaptation is largely preserved, and with more gesture repetitions can even improve. Replay anchors updates to previously observed exemplars, mitigating over-specialization and making adaptation more stable under limited or unbalanced target data. Notably, statistical alignment yields meaningful gains with very little target data (down to a small fraction of the available repetitions), suggesting that part of session variability can be compensated by adapting representations through statistics that are less sensitive to gesture-specific idiosyncrasies. Ablations that remove the replay buffer further isolate this effect: without replay, alignment still improves the adapted session with limited data, but retention on non-adapted sessions degrades, underscoring the role of replay in preventing forgetting. The price for this stability is higher computational cost than \ac{BN}, since adaptation requires gradient-based updates (albeit restricted to a small parameter subset).

Finally, meta-learning provides the expected upper bound when sparse labels are available at test-time. With a few calibration repetitions, MAML rapidly improves performance and can approach the accuracy of a fully supervised pipeline. This benefit, however, comes with trade-offs:  it is a supervised method; MAML requires a larger number of samples before delivering clear gains compared to statistical alignment; adaptation tends to produce more session-specific solutions, leading to a drop in terms of cross session impact. Augmenting MAML with an experience replay buffer mitigates this deterioration, albeit typically at the cost of reduced adaptation strength on the current session. For both MAML and statistical alignment, we use \ac{LoRA} to constrain the adaptation space for efficiency; moreover, updating only low-rank parameters makes it straightforward to revert the model to its base configuration when adaptation is not desired.

\begin{figure*}[ht!]
\centering
\includegraphics[width=0.95\textwidth]{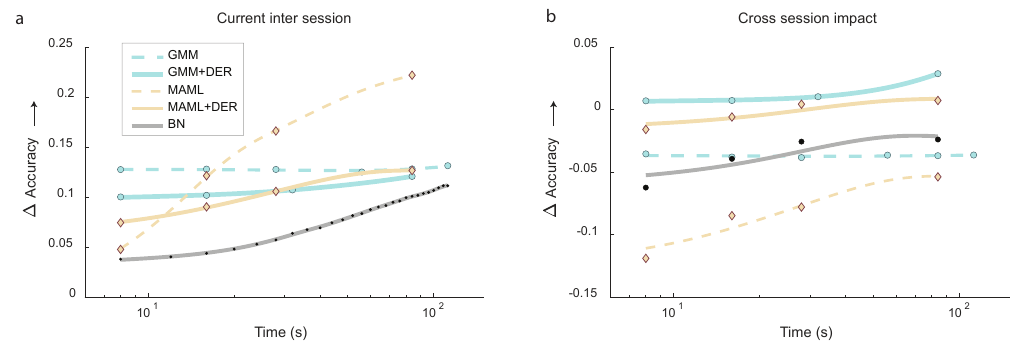}
\caption{Relative accuracy improvement of each test-time adaptation approach over the baseline with respect to the adaptive prefix $\mathcal{D}_{\tau}^{\leq t}$, measured in seconds rather than repetitions. Notice that a repetition lasts $4.5 \ s$ on average.}
\label{fig:comparative_study}
\end{figure*}

Figure~\ref{fig:comparative_study} summarizes the comparative trade-offs among the adaptation strategies. \ac{AdaBN} recovers performance on a slower timescale (Fig.~\ref{fig:comparative_study}a), for values of the adaptation speed $\beta$ that avoid overly aggressive updates; the black curve reports the average trend over the set of plausible $\beta$ choices of Fig.~\ref{fig:adaptation_results}, panels d and e. Statistical alignment methods are particularly effective in the low-data regime, in some cases surpassing MAML when the number of repetitions is smaller than the number of gesture classes, whereas MAML becomes more advantageous as additional labelled repetitions are provided. Figure~\ref{fig:comparative_study}b further reports performance on sessions held out from the current adaptation, showing that only replay-regularized methods (i.e. approaches that incorporate continual-learning principles) can consistently preserve, and even improve when considering more data, accuracy on non-adapted sessions, especially as more repetitions are observed.

Overall, all approaches substantially improve cross-session robustness compared to the non-adapted baseline, but with complementary strengths: \ac{BN} provides lightweight recalibration directly on the target stream; replay-regularized statistical alignment promotes stability and mitigates forgetting across sessions; and meta-learning further improves accuracy when supervisory signals are available during calibration. These results support the conclusion that lightweight test-time adaptation can significantly enhance inter-session generalization, an aspect that is often obscured by evaluations dominated by intra-session validation.



Although session shifts in \ac{EMG} are often attributable to practical factors such as sensor displacement and day-to-day re-donning, it remains valuable to equip the model with mechanisms that can autonomously detect and track distributional changes. The behavior of our causal \ac{BN} variant across different values of $\beta$, with no single setting performing best at all times, suggests that adaptation over multiple timescales could further improve robustness. Moreover, discrepancies between stored source statistics and test-time estimates computed at different timescales may provide a simple signal for monitoring out-of-distribution (OOD) conditions. Such a signal could be used to drive adaptive control of the deployment pipeline, for example, deciding when to increase or slow down adaptation, trigger a stronger replay-regularized update or a brief user calibration, or fall back to the base trained configuration when online updates appear unreliable.

In this regard, our results indeed suggest that different adaptation mechanisms could be combined to balance responsiveness and stability during deployment. In a practical wearable or embedded implementation, \ac{BN} adaptation could run continuously on-device to provide immediate, low-cost recalibration to changing conditions. In parallel, replay-regularized fine-tuning, via statistical alignment or via meta-learning when a brief user calibration is available, could be performed intermittently (e.g. off-device) to consolidate learning and update model parameters more robustly. Over time, such updates could leverage an expanding user-specific history of recordings, enabling progressively better personalization.

This hybrid pipeline would exploit complementary strengths, fast on-device adaptation and longer-term robustness, supporting scalable and energy-efficient \ac{EMG} decoding in real-world applications. As future work, this direction could be integrated with privacy, preserving personalization schemes (e.g. federated learning), where model updates are aggregated across users without centralizing raw \ac{EMG} data.

\section{Conclusions}
\label{sec:conclusions}

We investigated lightweight test-time adaptation for inter-session \ac{EMG}-based gesture recognition. While the baseline
achieved strong intra-session accuracy, performance degraded markedly on unseen sessions, reflecting the pervasive
out-of-distribution (OOD) shifts encountered in real-world neuromuscular sensing.

More broadly, these findings support the view that robustness in neural networks need not rely solely on training-time
coverage of all operating conditions, an assumption that can be particularly unrealistic for medical data and costly for
energy- and memory-constrained wearables. Instead, modest adaptation at deployment can correct session-specific shifts
using only the data observed in operation.

We evaluated three complementary strategies: causal \ac{AdaBN} updates of normalization statistics, replay-regularized
statistical alignment with \ac{LoRA} parameters, and few-shot meta-learning to mimic supervised calibration. All methods
improved inter-session robustness over the non-adapted baseline. \ac{AdaBN} provides the lowest-cost recalibration but
benefits from longer test-time horizons, whereas replay-regularized alignment yields the most stable behavior under
limited data by mitigating over-specialization and forgetting; meta-learning delivers the fastest gains when sparse labels
are available. Overall, minimal test-time adaptation offers a practical route to more reliable long-term \ac{EMG} decoding
for wearable and prosthetic control.

\printbibliography
\clearpage

\end{document}